\documentclass[runningheads]{llncs}
\usepackage[T1]{fontenc}
\usepackage{graphicx}
\usepackage[backend=bibtex, citestyle=ieee]{biblatex}
\usepackage{graphicx}

\usepackage{tabularx}
\usepackage{adjustbox}
\usepackage[table]{}
\usepackage{caption}
\usepackage{graphicx}
\usepackage{textcomp}
\usepackage{float}
\usepackage{array}
\usepackage{multirow}
\newcolumntype{P}[1]{>{\raggedright\arraybackslash}p{#1}}
\newcolumntype{C}[1]{>{\centering\arraybackslash}p{#1}}
\usepackage{hyperref}
\usepackage{url}
\hypersetup{
    colorlinks=true,
    linkcolor=black,      % Color for internal links (e.g., \ref, \cite)
    citecolor=black,      % Color for citations
    urlcolor=black        % Color for URLs
}
\begin{document}
\title{Facilitating the Emergence of Assistive Robots to Support Frailty: Psychosocial and Environmental Realities}
\titlerunning{Facilitating the Emergence of Assistive Robots to Support Frailty}
\author{Angela Higgins\inst{1} \and
Stephen Potter\inst{2} \and Mauro Dragone\inst{3} \and Mark Hawley \inst{2} \and Farshid Amirabdollahian\inst{4} \and Alessandro Di Nuovo\inst{5}
\and Praminda Caleb-Solly\inst{1}}
\authorrunning{A. Higgins et al.}
\institute{%
University of Nottingham, School of Computer Science, Nottingham, UK \email{angela.higgins@nottingham.ac.uk} \and University of Sheffield, Sheffield, UK \and 
University of Hertfordshire, Hatfield, UK \and 
Heriot-Watt University, Edinburgh, UK \and Sheffield Hallam University, Sheffield, UK
}
\maketitle              % typeset the header of the contribution
\abstract{While assistive robots have much potential to help older people with frailty-related needs, there are few in use. There is a gap between what is developed in laboratories and what would be viable in real-world contexts. Through a series of co-design workshops (61 participants across 7 sessions) including those with lived experience of frailty, their carers, and healthcare professionals, we gained a deeper understanding of everyday issues concerning the place of new technologies in their lives. A persona-based approach surfaced emotional, social, and psychological issues. Any assistive solution must be developed in the context of this complex interplay of psychosocial and environmental factors. Our findings, presented as design requirements in direct relation to frailty, can help promote design thinking that addresses people’s needs in a more pragmatic way to move assistive robotics closer to real-world use.\footnote{Author’s Accepted Manuscript. Released under the Creative Commons license: Attribution 4.0 International (CC BY 4.0) https://creativecommons.org/licenses/by/4.0/}

}

\section{Introduction}

Frailty can be characterised as the loss of resilience due to the adverse effects of ageing. It has physical, emotional and cognitive dimensions, which can hinder effective functioning in daily life \cite{fried2001frailty}, with effects influenced by social, environmental and economic circumstances \cite{lang2008neighborhood}. As populations worldwide age and countries experience care staff shortages, robots are often suggested as a means to support independent and healthy ageing. Designs which fail to take into account people's needs, wants, anxieties and circumstances are likely to lead to unacceptable, impractical, or ineffective robots. Currently, there are few deployable assistive robots on the market, and the research focus on using off-the-shelf robotic platforms has not been successful \cite{wright2023robots}. The primary aim of the Emergence Network \cite{EmergenceRobotics} was to identify and address the barriers to moving assistive robots from laboratories into people's homes. Here we present our findings from workshops that gathered requirements to guide the assistive robotics community towards developments that directly address frailty. 

\section{Related Work}

The ACCRA project used co-design to develop applications for the diagnosis of sarcopenia \cite{fiorini2019robot} and to support greater mobility \cite{fiorini2020co}. However, these investigations were constrained by the use of pre-existing robots \cite{fiorini2019robot}. More recently, the City4Age project created a co-design toolkit to develop interventions for older people. In their critique of other projects, they note that co-design often begins only once the robot has already been developed or chosen, skipping the initial, scoping design work and thereby creating an immediate disconnect between under-defined problem and over-specified robot ``solution'' \cite{bardaro2022robots}. With this in mind, one exercise in co-design of robots to encourage physical activity purposely ignored current robot capabilities and instead identified design guidelines in collaboration with occupational therapists and older people themselves \cite{antony2023co}. The resulting diversity of opinions highlighted the need for personalisation and adaptability, and for multimodal communication, ease of use, maintaining privacy, and behaving respectfully. The research described here takes a similar approach, seeking to identify interventions for frailty without committing to any pre-existing robotic platform, but rather with a commitment to exploring ideas suggested by participants, regardless of their current feasibility. Another multidisciplinary exercise in this area laid bare the extent of needs in this space, with over 70 distinct requirements initially identified \cite{garcia2018inclusion}. Here we go a step further by viewing needs through the prism of living with frailty, described through a list of functional and non-functional requirements. 

\section{Methods}

Inspired by the \emph{Double Diamond} model of design thinking \cite{thedesigncouncilDoubleDiamondDesign}, Emergence held co-design workshops corresponding to the \emph{"discover"} and \emph{"define"} phases of the first diamond. (Subsequent activities, not covered here, would take the results into the \emph{"develop"} and \emph{"deliver"} phases of the second diamond.)

Workshops during the \emph{"discover"} phase discussed issues surrounding activities of daily living (ADLs), self-management of frailty (SM), and the role of healthcare professionals (HCPs). Subsequent workshops were held to further \emph{"define"} the constraints on assistive robots. Research approval was obtained from the University of Nottingham research ethics committee, ref no. CS-2021-R40. 

The workshops, each lasting 3-4 hours, were held during 2022 around the UK, at times and places convenient for attendees. Recruitment took place through sector partners, professional networks, and word of mouth. Table \ref{table:workshops} summarises participation in the  workshops. A total of 61 participants were recorded, including some people attending multiple sessions. There were paired \textit{``define''} and \textit{``discover''} workshops at each location, with participants invited to attend both; however, some were able to attend only one. The second workshop took place around a fortnight later. All workshops were audio recorded, with manual notes taken. A graphic facilitator helped capture the content of several of the workshops through illustrations \cite{espiner2016innovation}. 

\textbf{Discover} The ADL, SM, and HCP discover workshops used personas to prompt discussion of the challenges and opportunities faced by people living with frailty. ADL workshops asked participants to identify and prioritise difficulties and opportunities during common daily activities. The SM workshop explored diet, exercise, mental health, sleep, and personal security. The HCP workshops reflected on key challenges faced by individuals during the patient journey and opportunities for robotic support.

\begin{table}[t]
\vspace{-0.6cm}
\caption{Workshop Information}
\centering
\begin{tabular}{|p{0.325\linewidth} | p{0.575\linewidth}|} 
 \hline
  \bf Stage & \bf Participants \\ \hline
    Discover: ADL1 & 8 older people, 1 housing support officer \\\hline
    Discover: SM1 & 8 older people, 2 OTs, 1 telecare rep, 1 carer \\\hline
    Define: ADL2 & 5 older people, 2 housing support officers \\\hline
    Define: SM2 & 8 older people, 3 HCPs \\\hline
    Discover: HCP1 & 4 HCPs \\\hline
    Discover: HCP2 & 10 HCPs \\\hline
    Discover+Define: ADL3 & 4 older people, 4 carers \\\hline    
\end{tabular}
\label{table:workshops}
\vspace{-0.6cm}
\end{table}
\textbf{Interim data analysis} A rapid analysis of the first workshops identified 6 prominent recurring everyday difficulties. Giving free rein to their imaginations, and unconstrained by questions of technical feasibility, the research team came up with the \textit{``speculative''} designs for a robot to address each. Each robot design was summarised in terms of the need/want it meets, its functioning, and its application or operation. These summaries, now incorporating an illustration from the graphic facilitator, were used in the \textit{"define"} workshops.

\textbf{Define} Workshops were held for both the ADL workshop groups and the SM group, and in each case the format was the same. Participants were presented with each speculative robot design and asked what they did or did not like about it, or what else they would need to know about it. The group was then asked for their opinions about various aspects of the robot and how they imagined it would behave. Specifically, they were asked to consider the robot's appearance, control, performance, practicalities, and concerns.

\textbf{Data Analysis} Collected data were subjected to a thematic analysis by two of the research team independently and then compared, with discrepancies resolved through discussion. A combined inductive and deductive approach was used, with deductive codes based on workshop topics and inductive codes developed from workshop discussions. These were sorted into functional and non-functional requirements as per the Volere specification template \cite{robertson2012mastering}. 

\section{Results and Discussion}

\subsection{Discover}

\textbf{ADL1 Workshop} Participants indicated that lack of motivation had a negative impact on many important ADLs (\textbf{n}umber of participants=4). They acknowledged that this can lead to further deterioration in mental and physical health, leading to a vicious circle of decline, which can be difficult to break once initiated. Some attributed problems with motivation to isolation, stating that well-being derived from \textit{“being cared about and car[ing] about yourself”}. Pets were considered a potential source of both motivation and companionship (n=3), but could be impractical due to housing restrictions or care demands. Fear of falling was identified as a barrier to performing ADLs (n=3), this was also a cyclical process, with anxiety leading to reduced activity, and to increased fear of falls. Another issue was the difficulty of remembering to take medication at the correct time (n=2). Practical and cost concerns were frequently raised (n=6), particularly the amount of electricity consumed by a robot (n=2). Older adults had often downsized their homes (n=2) and were wary of the amount of space that would be taken up by multiple robots.

\textbf{SM1 Workshop} Participants discussed already using technology to set reminders and for home security (n=3). However, they were apprehensive about technology leaving them open to scammers or the loss of the human touch. Frustration was expressed at the general lack of support for maintaining their health, particularly mental health care (n=3). Personalisation of a robot to the user and their needs was seen as a necessity, with control in the hands of the service user (n=3). As in the ADL workshops participants were receptive to the idea of robotic pets (n=10), and concerned about the practicalities of robot ownership (n=4). Communal robots were viewed as a possible solution, particularly for cleaning robots (n=4). 

\textbf{HCP1 and HCP2 Workshops} In HCP workshops all participants repeatedly returned to the theme of motivation and a need for self-efficacy. Due to typical multi-morbidity and the link between physical and emotional health, a holistic approach was emphasised (n=5). HCPs expressed frustration with current electronic healthcare records, and lack of data sharing, leaving them with insufficient information (n=7). They stated that they only ever saw a ``snapshot" of a patient, often at their worst after an adverse health event, therefore monitoring and assessment after a hospital stay may be enhanced by technology.

Suggested areas for robotic support included fall prevention (n=4), medication management (n=4), incontinence (n=3), and preventive interventions for pre-frailty (n=4). In general, technology that supports pre-frail individuals could benefit the wider care system (n=8). This was combined with acknowledgement of care workers whose job is often demanding and poorly paid (n=4). Technologies which would make care workers' jobs easier would help alleviate this burden (as would higher wages). 

\textbf{ADL3 Workshop} The ADL3 workshop consisted of both ``discover'' and ``define'' sessions and did not inform the speculative robot concepts. Themes from earlier workshops were echoed, including space constraints (n=2), fall anxiety (n=1), medication management (n=3), and interest in robotic pets (n=4). Participants expressed the need for both physical and emotional support after a fall or other medical incident, but expressed that equipment was often over-medicalised (n=3). Technology usability was raised, with participants stating that they particularly struggled with initial set up (n=3). This particular group expressed a desire to learn technological skills, noting the need for instructions to be tailored to people's accessibility issues.

\subsection{Define}

All speculative robot scenarios that were discussed and evaluated in ``define'' are available online \cite{EmergenceResources}, with short descriptions shown in Table \ref{tab:robot_descriptions}. 

\begin{table}[t]
\vspace{-0.6cm}
\centering
\caption{Descriptions of speculative robot scenarios}
\begin{tabularx}{\columnwidth}{|l|X|}
\hline
\textbf{Robot}    & \textbf{Description}                                                          \\ \hline
Motibot           & Detects low mood/lack of activity \& suggests activities               \\ \hline
Foodee            & Suggests recipes \& helps users follow them step-by-step             \\ \hline
EasyUp            & Offers a reassuring arm as you climb up and down stairs                        \\ \hline
AutoReach         & Hoovers, dusts and washes difficult-to-reach places                         \\ \hline
RoPet             & Robot pet companion, available in different models     \\ \hline
Toilittle         & Toileting robot with options according to user's situation \\ \hline
\end{tabularx}
\label{tab:robot_descriptions}
\vspace{-0.6cm}
\end{table}

\textbf{Personalisation} The need for robots to adapt to users’ physical, emotional, or practical context was considered essential for all scenarios by all define workshops (n=14). Examples from participants include that Motibot should tailor its motivational approach, EasyUp must match mobility needs, and Foodee should reflect personal taste and dietary requirements.

\textbf{Appearance} Robotic aesthetics would need to be adjusted depending on both the function and user of the robot (n=6). For example, RoPet should be soft and cuddly, Foodee should appear functional, and Toilittle should be discreet. However, participants felt that all robots should look unintimidating, trustworthy, and unobtrusive (n=4) but easy to spot to avoid collisions (n=3). 

\textbf{Interaction} Voice interaction was preferred by all older adults, sometimes supplemented with screens (n=3), but robots would need to understand regional speech and terminology. However, it was highlighted that this would be unsuitable for people with speech or hearing impairments.

\textbf{Maintenance} Participants wanted low-maintenance robots, with repair services readily available (n=4). This is clear in the case of AutoReach; a cleaning robot that would require regular cleaning would be self-defeating. However, some, like RoPet, could require some light upkeep, treating it more like a pet.

\textbf{Monitoring and Autonomy} Robotic monitoring was acceptable if tailored to the user's needs and frailty (n=7). Likewise, levels of robot autonomy might need to change depending on the user's day-to-day needs: for example, Foodee taking more initiative in food preparation if the user was unwell. Robots would also need to understand multiple steps within a process, such as not only reminding a user to take their medication but also confirming this had been done. 

\textbf{Safety and Security} Participants accepted that failures would occur with any technology, but highlighted that they should \textit{``fail safely''} (n=4), giving the example that EasyUp could call for help in case of failure. Trip hazards were also frequently mentioned as a potential hazard presented by domestic robots, so they should be highly visible and able to navigate to safe locations. Data privacy and protection from scams were also strongly emphasised.

\textbf{Human Touch} Robots for domestic tasks were considered useful (n=12) and may outperform humans in certain tasks. Robots like Toilittle were seen as beneficial for tasks that cause discomfort for both carers and service users. However, many indicated they would still prefer the “human touch” (n=3).

\textbf{Operating Environment} Robots must work in small, cluttered, real-world spaces, and not require home modifications (n=9). Those living in a residential setting noted that a robot like EasyUp could be shared between people (n=3). 

\textbf{Multipurpose} Participants doubted the viability of a single robot to do everything, while space considerations limit the desirability of single-task robots (n=3). There were suggestions on how each robot in the scenarios could be for both emotional and physical support, such as EasyUp also providing motivation.

\textbf{Speculative robots} Motibot, Foodee, EasyUp, and AutoReach were well received, with all participants seeing their value. Opinions on RoPet varied, with 3 major objections. Toilittle was a sensitive topic, which nobody wanted to use, most agreed it could be useful for those with greater health needs.
\begin{table*}[t]
\renewcommand{\arraystretch}{0.99}
\vspace{-0.6cm}
\setlength{\tabcolsep}{6pt}
\caption{Functional Requirements in Relation to Frailty}
\begin{tabular}{|p{0.3\linewidth}|p{0.64\linewidth}|}
\hline
\textbf{Requirement}                       & \textbf{In Relation to Frailty}                                                                                                                                                             \\ \hline
Support a holistic model of healthy ageing & Influenced by biological, psychological, social, and environmental factors                                                                                     \\ \hline
Address multiple facets of frailty         & People living with frailty often have multiple long-term conditions                                 \\ \hline
Reduce isolation                           & Isolation can impact health, so a robot companion should also encourage human-to-human interaction                                                        \\ \hline
Not replace or undermine human care        & Enable the care of frail older people to be done with dignity and improve jobs within the care sector                                                                \\ \hline
Be affordable                              & Older adults often have less disposable income, so robots should provide good value for money                                                                                                   \\ \hline
Be part of a service ecosystem             & The whole service delivery model should be considered, including who buys and distributes the robots              \\ \hline
Have adjustable levels of autonomy         & Over-reliance may increase frailty, but when users need extra support, the robot should be able to provide it   \\ \hline
Have adjustable levels of monitoring       & This may not be needed for all, but this may change if their condition progresses or after a health incident      \\ \hline
\end{tabular}
\label{tab:func_requirements}
\vspace{-0.6cm}
\end{table*}

 \subsection{Requirements}
The problems, opportunities and ideas from our participants are shown as functional (Table \ref{tab:func_requirements}) and non-functional (Table \ref{tab:non_func_requirements}) requirements. These have been related to specific needs for people living with frailty.

\section{Conclusion}

Findings suggest that older people and healthcare professionals are willing to use robotic interventions to support independence and manage frailty. However, despite an abundance of research into assistive robots, few interventions have made the leap into the real world. As others have noted, researchers often start with a specific robotic platform in mind (and preconceived notions of the needs and abilities of older people), rather than seeking basic requirements. With this research, we hope to redress the balance and guide the development of assistive robots that are useful, usable, acceptable, and feasible.

\begin{table}[t]
\renewcommand{\arraystretch}{0.99}
\vspace{-0.6cm}
\setlength{\tabcolsep}{6pt}
\caption{Non-Functional Requirements in Relation to Frailty}
\label{tab:non_func_requirements}
\begin{tabular}{|p{0.16\linewidth}|p{0.32\linewidth}|p{0.42\linewidth}|}
\hline
\textbf{Category} & \textbf{Requirement} & \textbf{In Relation to Frailty} \\ \hline

\multirow{3}{*}{\parbox[t]{\linewidth}{\centering\textbf{Look and Feel}}}
& Highly visible & Avoid creating trip hazards \\ \cline{2-3}
& Unobtrusive appearance & No distress for those with cognitive impairments \\ \cline{2-3}
& Non-medical & Avoid medicalisation in daily life \\ \hline

\multirow{5}{*}{\parbox[t]{\linewidth}{\centering\textbf{Usability \& Humanity}}}
& Easy for older people to use & Support users when they are alone \\ \cline{2-3}
& Easy for carers, HCPs and others to use & Allow people in the user’s wider support circle to use it \\ \cline{2-3}
& Reduce cognitive load & To enhance memory \& concentration \\ \cline{2-3}
& Personalisation & Suit the user's changing needs \\ \cline{2-3}
& Multimodal control & Interact with users with multiple, intersecting accessibility needs \\ \hline

\multirow{2}{*}{\parbox[t]{\linewidth}{\centering\textbf{Performance}}}
& Fail safely & Prevent injury or stranding users \\ \cline{2-3}
& Awareness of home hazards & Does not create more hazards \\ \hline

\multirow{3}{*}{\parbox[t]{\linewidth}{\centering\textbf{Operational \& Environment}}}
& Suitable for small homes & Older adults often downsize \\ \cline{2-3}
& Consider shared environments & More older people live in shared or sheltered accommodation \\ \cline{2-3}
& Fit to the environment & People do not want to change their homes for robots \\ \hline

\multirow{2}{*}{\parbox[t]{\linewidth}{\centering\textbf{Maintenance \& Support}}}
& Full maintenance services & Older people are unlikely to be able to maintain a robot \\ \cline{2-3}
& Encourage caring for the robot & Build a sense of responsibility for day-to-day maintenance \\ \hline

\parbox[t]{\linewidth}{\centering\textbf{Security}}
& Controlled by primary user & Protection from bad actors \\ \hline

\multirow{2}{*}{\parbox[t]{\linewidth}{\centering\textbf{Cultural}}}
& Understand regional difference & Users do not have to alter their language for the robot \\ \cline{2-3}
& Different language options & To negate language barriers \\ \hline

\end{tabular}
\vspace{-0.6cm}
\end{table}

Older people, their carers, and healthcare professionals shared their experiences of living with frailty, and opportunities to develop technological interventions. Cognitive, emotional and physical challenges were identified related to daily life and self-management, and all groups emphasised the need for a holistic approach to managing frailty. Technology needs to be personalised, as older people are not a homogeneous group. Safety, trust, and user-friendliness are key issues, and people are concerned about the cost and logistics of robotic provision and support, so developers should consider the design and cost-effectiveness of the wider service model. Finally, we have also created a set of ``empathy cards'' using the illustrations made during the workshops \cite{EmergenceResources}. To aid dissemination, we use these as prompts and provocations in workshops with developers and other stakeholders to help promote more holistic design concepts for assistive robots. 

\section{Limitations}

The study involved a limited, self-selecting sample, so findings may not generalise to the wider population. Moreover, the findings are specific to a particular cohort, time, and place. At this stage, we excluded participants with technical backgrounds to allow future phases of the project to engage developers, incorporating insights from older adults and HCPs into robotic design.

\section*{ACKNOWLEDGMENTS}

The authors thank all the participants, as well as Rebekah Moore and Sam Church. This work was funded by the EPSRC, UK [Grant numbers  EP/W000741/1 and EP/S023305/1] and the Horizon Centre for Doctoral Training.

\printbibliography
\end{document}